\documentclass[lettersize,journal]{IEEEtran}
\pdfoutput=1
\usepackage{amsmath,amsfonts}
\usepackage{algorithmic}
\usepackage{algorithm}
\usepackage{array}
\usepackage[caption=false,font=normalsize,labelfont=sf,textfont=sf]{subfig}
\usepackage{textcomp}
\usepackage{stfloats}
\usepackage{url}
\usepackage{verbatim}
\usepackage{graphicx}
\usepackage{cite}
\usepackage{multirow}
\usepackage{siunitx}
\usepackage{booktabs}
\usepackage{caption}

\begin{document}

\title{Voxelized 3D Feature Aggregation for Multiview Detection}


\author{Jiahao Ma, Jinguang Tong, Shan Wang, Zicheng Duan, Chuong Nguyen

\thanks{This paper was produced by the IEEE Publication Technology Group. They are in Piscataway, NJ.}
\thanks{Manuscript received April 19, 2021; revised August 16, 2021.}}

\markboth{Journal of \LaTeX\ Class Files,~Vol.~14, No.~8, August~2021}%
{Shell \MakeLowercase{\textit{et al.}}: A Sample Article Using IEEEtran.cls for IEEE Journals}

\IEEEpubid{0000--0000/00\$00.00~\copyright~2021 IEEE}

\maketitle

\begin{abstract}

Multi-view detection incorporates multiple camera views to alleviate occlusion in crowded scenes, where the state-of-the-art approaches adopt homography transformations to project multi-view features to the ground plane. However, we find that simple 2D transformations do not take into account the object's height, leading to ground plane features.
To solve this problem, we propose VFA, short for Voxelized 3D Feature Aggregation, for feature transformation and aggregation for multi-view detection. Specifically, we voxelize the 3D space, project the voxels onto each camera view, and associate 2D features with these projected voxels. This allows us to identify and then aggregate 2D features along the same vertical line, alleviating projection distortions to a large extent. 
Additionally, because different kinds of objects (human vs. cattle) have different shapes on the ground plane, we introduce the oriented Gaussian encoding to match such shapes, leading to increased accuracy and efficiency. 
We perform experiments on multiview 2D detection and multiview 3D detection problems. Results on four datasets (including a newly introduced MultiviewC dataset) show that our system is very competitive compared with the state-of-the-art approaches. 
Code and MultiviewC are released at \url{https://github.com/Robert-Mar/VFA}.
\end{abstract}

\begin{IEEEkeywords}
Multiview detection, 3D object detection, synthetic multiview dataset.
\end{IEEEkeywords}

\section{Introduction}
\label{intro}
Multi-view 3D object detection\cite{2008Multicamera,8578626, haoran2021mvm3det} on synchronized frames from different calibrated cameras is considered as an effective solution to the occlusion problem faced by a single camera. A multi-view perception system leverages images that come from different perspective views with overlapping fields of view to complement each other to alleviate blind spots in the field of view. The essence of multi-view detection is how to effectively fuse features from multiple camera views. 
\IEEEpubidadjcol
State-of-the-art systems \cite{hou2020multiview,hou2021multiview,Song_2021_ICCV} in this field leverage camera poses and are typically based on 2D transformations, \emph{i.e.,} projecting multi-view image feature onto ground plane. With such transformations the feature projection procedure follows the geometric constraints of the cameras, so that the same point on the ground plane can to some extent aggregate features depicting the same object from multiple cameras. \IEEEpubidadjcol
\begin{figure}
\begin{center}
\includegraphics[width=0.99\columnwidth]{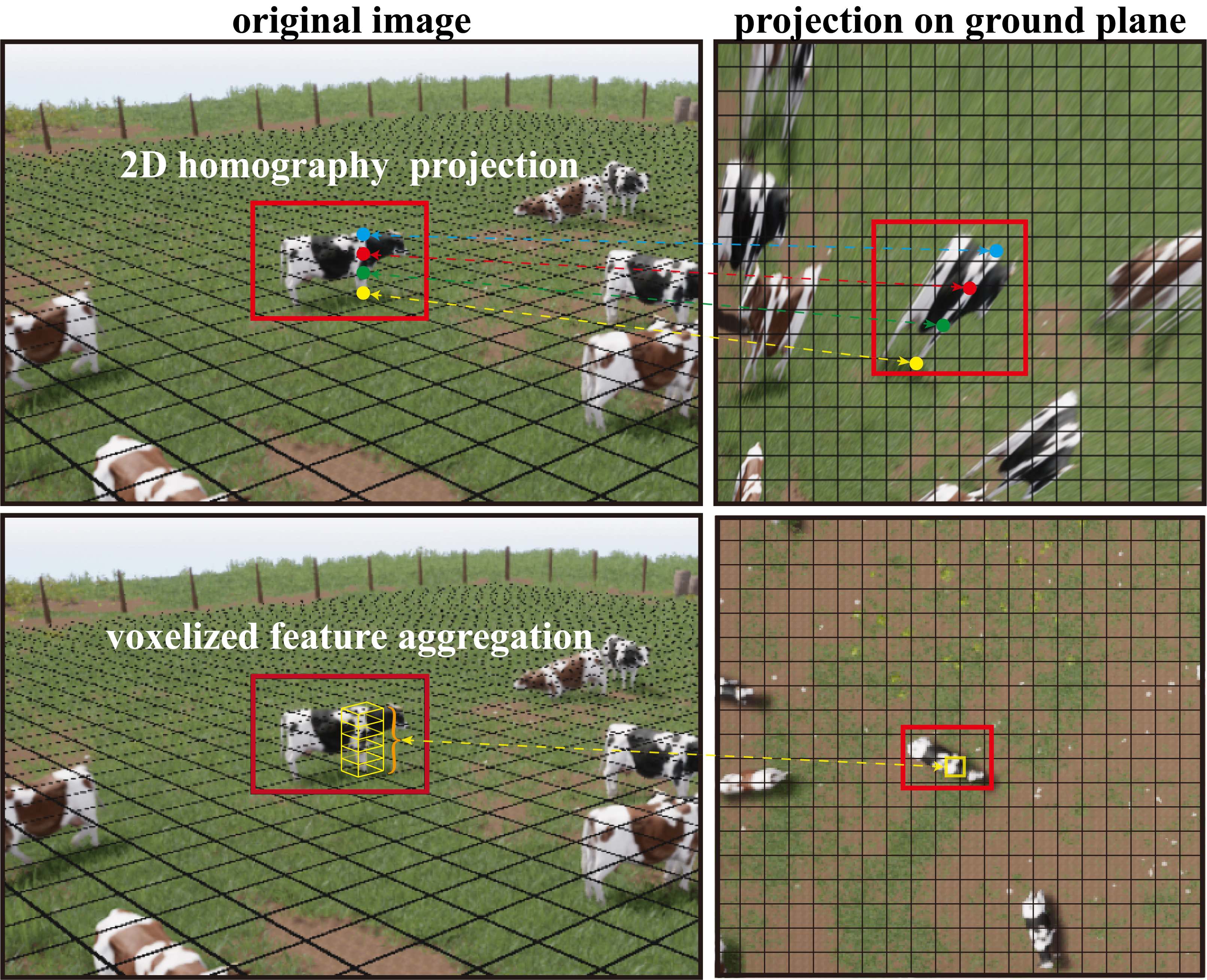}

\end{center}
\caption{
Illustration of 2D homography transformation \emph{vs.} voxelized 3D feature aggregation (VFA). $\textbf{Row 1}$: Features along the vertical direction of the same cattle are projected onto different ground plane positions by 2D homography transformation, resulting in projection distortions. $\textbf{Row 2}$: Through VFA, features along cattle's font leg are voxelized and accurately mapped to the same corresponding area on the ground plane, thus alleviating projection distortion. The bottom right bird's-eye-view image is generated by the simulation engine for  illustration purpose.
}
\label{homography_vs_VFA}
\end{figure}

\IEEEpubidadjcol
However, these 2D transformations do not consider the 3D scene geometry, and would make distorted projections of features along the $Z$ axis (vertical direction) of same object. That is to say, features of the same object at different height levels are not projected onto the same position of the ground plane. In effect, while a position on the ground plane is supposed to aggregate features of the same object at that position, it actually receives features from a mixture of objects, leading to the polluted features. We found that such features not only compromise the system accuracy, but also slow down the training process. In Fig. \ref{homography_vs_VFA}, we showcase an example where the homography 2D transformation mistakenly projects four colored points on a vertical line to different positions on the ground plane. Therefore, it is critical that our system is aware of such distortion to identify features that are on the same vertical direction in a multi-view system.
\IEEEpubidadjcol
To address the problem, we propose Voxelized 3D Feature Aggregation, or VFA, a 3D-aware projection method for multi-view object detection. In a nutshell, VFA explicitly considers the 3D clues by partitioning the 3D volumn-of-interest into cubic voxels. This voxelization process allows us to select the $Z$ axis in the 2D multi-view images, so that we can identify and aggregate features on the same vertical line. Our method has the following characteristics and advantages. First, different from existing 2D transformation approaches, our method allows each position on the ground plane to receive and aggregate 2D features that are of the same vertical direction, thus mostly depicting the same object. This endows the aggregated features with a less distorted pattern, thus improving both recognition accuracy and training efficiency. Second, the underlying prior knowledge of our method is that objects are usually bounded in a vertical volume starting from the ground, due to the effect of gravity. As such, our method in principle is applicable to various objects such as human and cattle as demonstrated in this work.  

\begin{figure*}[ht!]
\begin{center}
\includegraphics[width=0.95\textwidth]{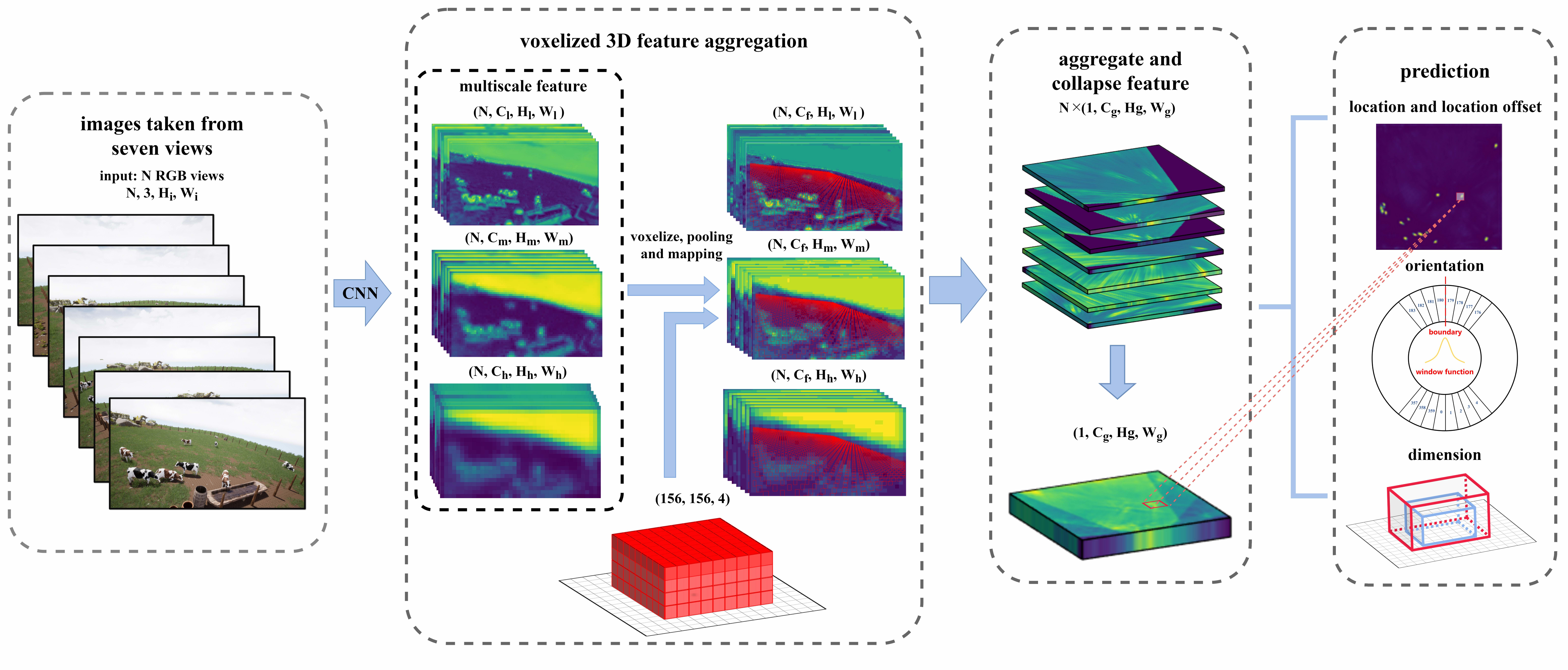}

\end{center}
\caption{
System workflow. Given multiview input images of shape $[3, H_i, W_i]$ from $N$ cameras, we use a shared convolutional neural network (CNN) backbone to extract multi-scale image features, denoted from low (``l''), medium (``m'') to high resolution (``h'') as $[N, C_l, H_l, W_l]$, $[N, C_m, H_m, W_m]$ and $[N, C_h, H_h, W_h]$, respectively. To perform voxelized 3D feature aggregation, we create a cuboid made of smaller voxels (red), where we use a grid size of $156\times156\times4$ as an example. We project each voxel to multi-scale features, and within each projected voxel, we extract and pool the corresponding 2D features across multiple views, so that 
we obtain $N$ stacks of feature maps of size $[1, C_g, H_g, W_g]$ coming from $N$ cameras, where subscript ``g'' means ``global''. We concatenate and collapse these feature maps along the channel dimension to be used for final 3D prediction.
Four output heads are used during prediction. A confidence map and an offset map are predicted to jointly localize objects. The other two heads are responsible for orientation and dimension estimation.
In this figure, $C, H, W$ represent the number of channels, image height and width, respectively. }
\label{fig:MOFT3D}
\end{figure*}

In addition, to deal with objects (\emph{e.g.,} cattle) that have an oriented shape after being projected to the ground plane, we propose a simple technique applying oriented Gaussian distribution (OGD) to encode the ground truth location and orientation on the ground plane. This minor contribution brings further improvement in detection accuracy and training efficiency for multi-view cattle detection. 

We perform extensive experiments on four datasets intended for multi-view detection, where we show our system yields very competitive accuracy compared with the state-of-the-art methods. Among the four datasets is a synthetic dataset for multi-view cattle detection, named MultiviewC, newly introduced by this paper.
The points made in this paper are summarized below. 
\begin{itemize}
\item Major: an approach VFA to allow $Z$-axis-aware feature extraction aggregation for multi-view detection.
\item Minor: an improved label embedding method, oriented Gaussian distribution, for directional objects (\emph{e.g., cattle}) detection.
\item A synthetic 3D multi-view dataset, MultiviewC, which includes cattle targets of diverse sizes and actions.
\end{itemize}



\section{Related Works}\label{related}
\setlength{\parskip}{0pt}
\textbf{Monocular camera based detection.}
Remarkable achievements have been made in 3D object detection over recent years. Due to the lack of depth information in the images, researchers have proposed fusion-based methods\cite{Chen_2017_CVPR, xu2018pointfusion, ku2018joint, liang2018deep}. For instance, MV3D\cite{Chen_2017_CVPR} encodes the sparse point cloud with multi-view representation and performs region-based feature fusion. Pointfusion\cite{xu2018pointfusion} uses two networks to process images and raw point cloud data separately, and fuses them at feature level. In addition to predicting depth by fusing point cloud, more approaches have been proposed to directly predict 3D Boxes using convolutional neural network. Some methods\cite{chen20153d, chen2016monocular} tend to perform 2D/3D matching via exhaustively sampling and scoring 3D proposals as representative templates. And some methods\cite{mousavian20173d, li2019gs3d, Li_2019_CVPR} start with accurate 2D bounding boxes directly to roughly estimate 3D pose from geometric properties obtained by empirical observation.
\\
\indent\textbf{Multiple views based detection.}
To detect object under heavy occlusion, several methods have been developed based on multiple perspectives. Fleuret \emph{et al.} \cite{2008Multicamera} proposes Probabilistic occupancy map to estimate the probabilities of occupancy using Multi-view streams jointly. To aggregate spatial neighbor feature, Conditional Random Field (CRF)\cite{roig2011conditional, Baqu2017DeepOR} has been exploited. Lima \textit{et al}.  \cite{Lima_2021_CVPR} present a multiview detection model without training but estimate the standing point within each 2D detection area and obtain spatial position of pedestrians via solving the clique cover problem. Yan \textit{et al}. \cite{PR} calculate the likelihood of pedestrian presence in each detection and clusters people positions via minimize a logic function. Hou \emph{et al.} \cite{hou2020multiview} adopts feature perspective transformation to aggregate multi-view data and regress pedestrian occupancy as a Gaussian distribution. Lima \emph{et al.} \cite{lima2021generalizable} presents a multi-camera detection model without any training that tends to integrate multiple views via graph-based method. Song \emph{et al.} \cite{Song_2021_ICCV} proposes SHOT to alleviate the projection errors by multi-height-level homography transformation.
\\
\indent\textbf{Inference in the bird's-eye-view frame.}
Models using intrinsics and extrinsics to tackle the difficult problem of predicting birds-eye-view representation received significant interest recently. A common approach\cite{ammar2019geometric, lin2012vision, hou2020multiview, haoran2021mvm3det, hou2021multiview, Song_2021_ICCV} is to use inverse perspective mapping (IPM) to map front-view image onto the ground plane via homography projection. OFT-Net \cite{roddick2018orthographic} projects a fixed volume of voxels onto multiview images to collect features and complete 3D detection on the bird's-eye-view feature representation. ``Lift, Split, Shoot" idea \cite{philion2020lift} is proposed to infer birds'-eye-view representation by lifting each image into a frustum of features and collapsing all frustum into a rasterized bird's-eye-view grid. These approaches mentioned above focus on autonomous driving scenarios where these sensors are closed with each other. Our proposed approach targets at multi-view detection where relative positions between sensors are far enough and different perspective views still have large overlapping fields of view. 

\section{Method}
This section details the proposed system that mainly deals with projection distortions when aggregating multi-view features. 
There are three components, including multi-scaled feature extraction (Section \ref{Multi-view Feature Extractor}), voxelized feature aggregation (Section \ref{VFA}) and  multi-branch estimation (Section \ref{RGD} and Section \ref{Orientation and Dimension}). 
\subsection{Feature Map Extraction}\label{Multi-view Feature Extractor}
In VFA, first, given N images of size [$H_i$, $W_i$] as input ($H_i$ and $W_i$ denote the image height and width respectively and N is determined by the number of cameras), the proposed network includes ResNet-18 \cite{He_2016_CVPR} as feature extractor to generate a hierarchy of multi-scale 2D feature maps from each input image. Multi-scale feature maps are generated separately for N input images, with shared weights for all calculation. After extracting features, a multi-scale feature representation will be generated by concatenation of same scale features to the dimensions of [N, $C_l$, $H_l$, $W_l$], [N, $C_m$, $H_m$, $W_m$], and [N, $C_h$, $H_h$, $W_h$] where the subscripts $l$, $m$ and $h$ denote low, middle and high resolution feature representation. $C_l$, $H_l$ and $W_l$ represent the channel, height and width of low resolution feature. Before voxel projection, we adjust the number of channels of different scaled feature to be consistent through a $1\times1$ convolution, with height and width remaining unchanged.
\begin{figure}
\begin{center}
\includegraphics[width=0.99\columnwidth]{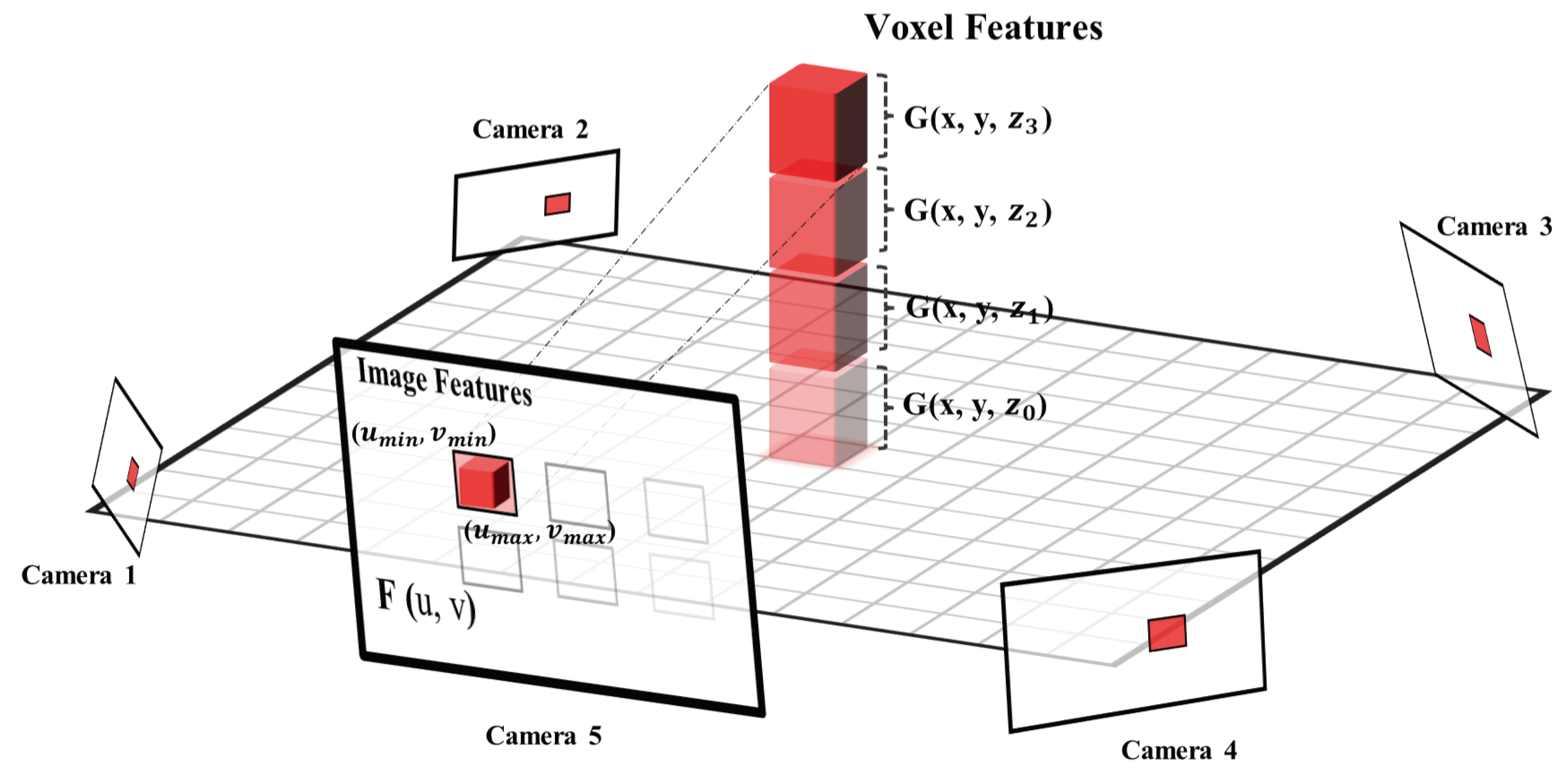}

\end{center}
\caption{
A closer look at voxelized 3D feature aggregation. The 3D voxels are created upon the ground plane and projected to each camera view. In each view, image features $\mathbf{F}(u, v)$ inside each 2D bounding box of projected voxels are obtained, aggregated and assigned to the corresponding 3D voxel feature $\mathbf{G}(x, y, z)$.
}\label{MOFT_fig}
\label{MOFT}
\end{figure}
\subsection{Voxelized Feature Aggregation}\label{VFA}
 The essence of VFA is to map multi-view image features $\mathbf{F}(u, v) \in \mathbb{R}^{n}$ to corresponding voxel grids, generating 3D voxel feature $\mathbf{G}(x, y, z) \in \mathbb{R}^{n}$. The voxel grids are defined beforehand, whose size depends on the proportion of movement range of all objects and grid height is determined by the height of the objects. To explore the impact of size of voxel grids on model performance, extensive experiments are conducted as described in Section \ref{voxel_grids_experiment}. For a given voxel grid location $(x, y, z) \in \mathbb{R}^{n}$, voxel feature  $\mathbf{G}(x, y, z)$ is corresponding to the pooling result of the 2D projection of voxel on image feature. The pooling area can be estimated as the bounding box of the voxel 2D projection. How VFA aggregates multi-view features can be summarized in the following steps:

\begin{enumerate}
    \item Before inference, a voxel grid is generated at location $(x, y, z) \in \mathbb{R}^{n}$ and with the size of $(l, w, h)$.
 
    \item Project eight corners of each voxel to multi-view image planes. Given a 3D world position $(x, y, z)$, we can calculate its projected image coordinate $(u, v)$ following,
        \begin{equation}
            \begin{split}
            \hspace{-8mm}
            \gamma\!\left(\!\begin{array}{l}
            \!u\! \\
            \!v\! \\
            \!1\!
            \end{array}\!\right)\!\!&=\!P_{\theta}\!\left(\!\begin{array}{l}
            \!x\! \\
            \!y\! \\
            \!z\! \\
            \!1\!
            \end{array}\!\right)\!\!=\!K[R \!\!\mid\! t]\!\!\left(\!\begin{array}{l}
            \!x\! \\
            \!y\! \\
            \!z\! \\
            \!1\!
            \end{array}\!\right)\!\!,
            \end{split}
        \end{equation}
   
    where $P_{\theta}$ denotes projection matrix, $K$ and $[R \!\!\mid\! t]$ denote intrinsic matrix and rotation-translation matrix respectively, and $\gamma$ is an arbitrary scale factor. For each voxel, we will get eight projected corners $c_i \left(u_i, v_i\right), i\in\{1, \ldots, 8\}$, on the image plane. 

    \item Calculate the bounding box with the top-left corner $(u_{min}, v_{min})$ and bottom-right corner $(u_{max}, v_{max})$ from 8 projected voxel corners.

    \item Extract the learned feature by average pooling over the bounding box of voxel 2D projection and assign to the corresponding location in the voxel feature $\mathbf{G}$:
        \begin{equation}
            \begin{split}
                \mathbf{G}(x, y, z)=\frac{\sum\limits_{u=u_{min}}^{u_{max}} \sum\limits_{v=v_{min}}^{v_{max}} \mathbf{F}(u, v)}{\left(u_{max}-u_{min}\right)\left(v_{max}-v_{min}\right)}.
            \end{split}
        \end{equation}
\end{enumerate}

As shown in Fig. \ref{MOFT_fig}, multi-view image features are projected to appropriate location in 3D voxel grids. Directly applying 3D Convolutional Neural Networks (3D CNNs) on compact 3D voxel feature representation will cost huge computational resources. Thus, before multi-branch predicting, the 3D voxel feature will be collapsed along the vertical axis, becoming 2D BEV features.

\textbf{Our difference from and advantage over \cite{Song_2021_ICCV}}. Both our method and \cite{Song_2021_ICCV} aim to address the projection distortion problem. To this end, Song \emph{et al.} \cite{Song_2021_ICCV} generate multiple ground planes at different heights onto which the 2D features are projected. This method, while alleviating overall distortions to some extent, its use of image interpolation in 2D projection does not properly sample the multiview features.
In contrast, by projecting voxels and pooling features inside the voxel bounding boxes, our method samples the multiview features properly according to the perspective distortion.
As to be shown in Fig. \ref{H_NUM_vs_Perf_FPS} A, we observe consistent improvement when implementing the VFA module on top of \cite{Song_2021_ICCV}. Another difference is our use of oriented Gaussian distribution to represent the probability of object location as described in the following section.

\subsection{Oriented Gaussian Distribution for Ground-truth Modeling}\label{RGD}
\begin{figure}
\begin{center}
\includegraphics[width=0.99\columnwidth]{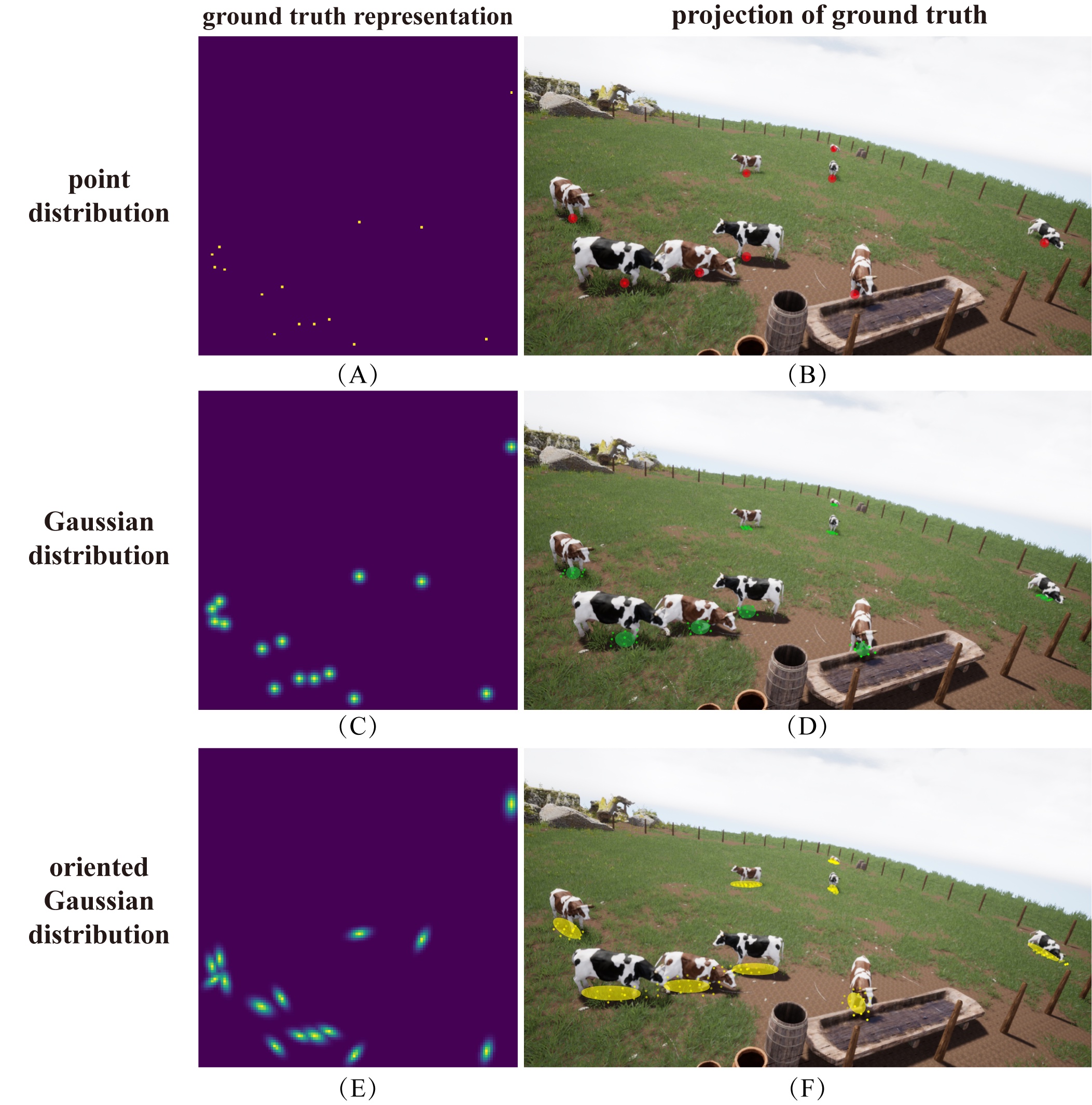}

\end{center}
\caption{
Comparison of three ground truth modeling methods. \textbf{Top}:
point distribution provides a discrete ground truths, making it more difficult for network to converge. \textbf{Middle}: spherical Gaussian modeling is not suitable for objects that have rectangle or oriented ground truth shapes. To adjust our system to objects like cattle, we design the oriented Gaussian modeling technique (\textbf{Bottom}), which encodes both the position and orientation of the projected ground truths. \textbf{B, D, F} show that projected oriented Gaussian distributions match better to the ground area under the cattle than Point and Gaussian distributions.
}\label{GT_Representation}
\end{figure}
Some methods \cite{2008Multicamera, Baqu2017DeepOR} propose a confidence map of probability as individual pixels to encode the positions of objects, as shown in Fig.\ \ref{GT_Representation} (\textbf{A}). Other methods \cite{hou2020multiview, liu2020training} introduce smooth Gaussian distribution to represent the region where objects are located, as in Fig. \ref{GT_Representation} (\textbf{C}). 
However we find it intuitive to adopt oriented Gaussian distribution to encode the elongated objects on confidence map. As shown in Fig. \ref{GT_Representation} (\textbf{E}), the center of oriented Gaussian distribution indicates the location of object on BEV plane. In addition, the size and rotation angle of oriented Gaussian distribution is relative to the size (width and length) and the orientation of object respectively.
Given a set of ground truth objects at $[x_i, y_i, z_0]^{T}$, (all objects are on the ground, so $z_0$ is 0 by default), with dimension $[l_i, w_i, h_i]^{T}$ and orientation $\beta$, where $i=1,\dots,N$, the oriented Gaussian distribution can be represented by:
    \begin{equation}
        \mathbf{S}(x, y)=\exp \left(-\frac{\left(x_{i}-x\right)^{2}}{2 \sigma_{l}}-\frac{\left(y_{i}-y\right)^{2}}{2 \sigma_{w}}\right),
    \end{equation}
where $\sigma_{l}= (l_i * \alpha)^{2}$, $\sigma_{w}= (w_i * \alpha)^{2}$, and $\alpha$ is set to 0.01 by default. In the light of orientation encoding, image rotation with bilinear interpolation \cite{smith1981bilinear} is applied to the region of oriented Gaussian distribution.
\begin{figure*}[ht!]
\begin{center}
\includegraphics[width=0.95\textwidth]{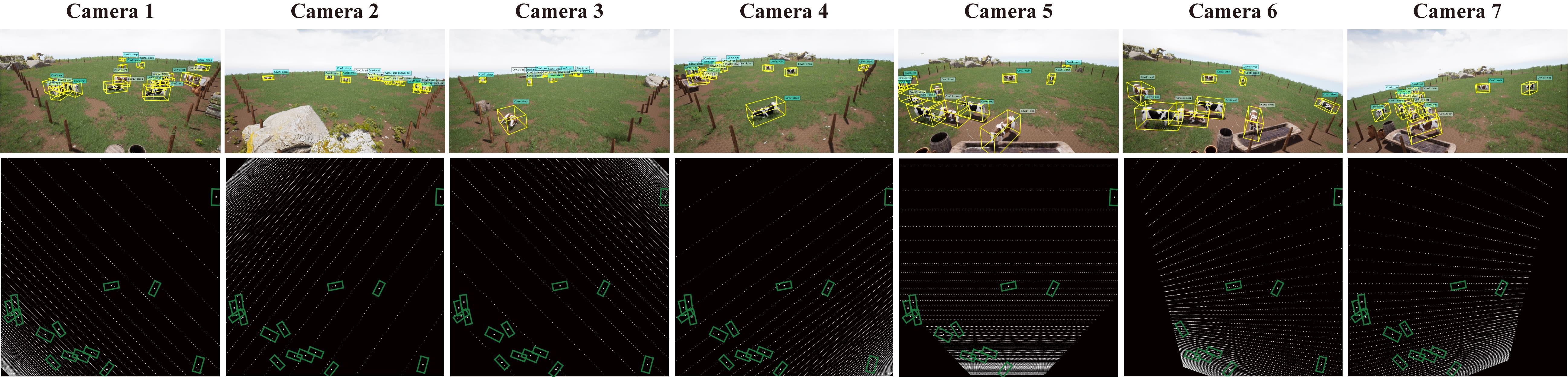}

\end{center}
\caption{
Sample images and their ground truths of the newly introduced  \textbf{MultiviewC} dataset, designed for multiview 3D detection of cattle. $\textbf{Row 1}$: images from 6 camera views with 3D bounding boxes and  cattle IDs, and action categories (the last two are not utilized in this paper).
$\textbf{Row 2}$: visualization of cattle bounding boxes on bird's-eye-view plane. 
}
\label{MultiviewC}
\end{figure*}
\subsection{3D Bounding Box Prediction}\label{Orientation and Dimension}
For 3D object detection, we need to predict the location, orientation and dimension of each object bounding box.
For location, we first predict a confidence map which is a reduced version of movement range of all objects. For instance, we design a confidence map of size $156 \times 156$ for MultiviewC dataset that covers a $39 m \times 39 m$ area.
Each point value on the confidence map indicates the possibility of object existence within its coverage.
When zooming in, lots of coordinate conversions fail to be divisible. Thus, we introduce an additional output head to predict relative position offset $\boldsymbol{\Delta}_{\operatorname{pos}}(x, y)$:
\begin{equation}
    \boldsymbol{\Delta}_{pos}(x, y)=\left[\begin{array}{ll}
    \frac{x_{i}}{\gamma} - \lfloor\frac{x_{i}}{\gamma}\rfloor  & \frac{y_{i}}{\gamma} - \lfloor\frac{y_{i}}{\gamma}\rfloor
    \end{array}\right]^{\top},
\end{equation}
where $\gamma$ is a scaling factor from original movement range to designed confidence map. In different tasks, $\gamma$ is different. For instance, in MultiviewC dataset, $\gamma = 3900 \div 156 = 25$.
To alleviate the class imbalance, the confidence map $\mathbf{S}(x, y)$ is trained via focal loss \cite{lin2017focal} to regress object occupancy on bird's-eye-view plane. In addition to location, orientation and dimension are also indispensable in 3D object detection. We therefore append two additional networks to predict the orientation and dimension. First, for the orientation head, different from previous works that propose multi-bins to classify the angle range and regress the angle offset, we introduce circle smooth label (CSL) \cite{yang2020arbitrary} to complete angle prediction. 
CSL expression is as follows:
\begin{equation}
    C S L(x)=\left\{\begin{array}{c}
            g(x), \theta-r<x<\theta+r \\
            \!\!\!\!\!0, \quad \text { otherwise }
            \end{array}\right.,
\end{equation}
where $g(x)$ is window function, $r$ is the radius of window function and $\theta$ is the angle that need to predict. In our task, we take Gaussian function as window function $g(x)$ with $r$ equal to $6$. Due to oriented Gaussian distribution being able to preserve the orientation cues, orientation head with CSL directly operates on bird's-eye-view plane, predicting the orientation of each grid cell on a confidence map.

When it comes to dimension head, we predict the scale offset $\boldsymbol{\Delta}_{\operatorname{dim}}(x, y)$, which is given by,
\begin{equation}
    \boldsymbol{\Delta}_{\operatorname{dim}}(x, y)=\left[\begin{array}{lll}
    \log \frac{l_{i}}{\bar{l}} & \log \frac{w_{i}}{\bar{w}} & \log \frac{h_{i}}{\bar{h}}
    \end{array}\right]^{\top},
\end{equation}
where $d_i = [l_{i} \ w_{i} \ h_{i}]$ denotes the dimension of ground truth object $i$ and $\bar{d} = [\bar{l} \ \bar{w} \ \bar{h}]$ represents the mean dimension over all objects.

\begin{table*}[ht!]
\begin{center}
    \small
    \resizebox{0.99\linewidth}{!}{
    \begin{tabular}{l cccc c cccc}
    \toprule
    \multirow{2}{*}{Method} & \multicolumn{4}{c}{ Wildtrack } & \quad & \multicolumn{4}{c}{ MultiviewX } \\
    \cmidrule[0.4pt]{2-5} \cmidrule[0.4pt]{7-10}
    & MODA & MODP & Precision & Recall & \quad & MODA & MODP & Precision & Recall \\
    \midrule[0.4pt]
    RCNN $\&$ clustering\cite{xu2016multi}     & $11.3$ & $18.4$ & $68$ & $43$ & \quad & $18.7$ & $46.4$ & $63.5$ & $43.9$ \\
    POM-CNN \cite{2008Multicamera}                        & $23.2$ & $30.5$ & $75$ & $55$ & \quad & $-$ & $-$ & $-$ & $-$ \\
    DeepMCD \cite{chavdarova2017deep}                        & $67.8$ & $64.2$ & $85$ & $82$ & \quad & $70.0$ & $73.0$ & $85.7$ & $83.3$ \\
    Deep-Occlusion \cite{Baqu2017DeepOR}                 & $74.1$ & $53.8$ & $95$ & $80$ & \quad & $75.2$ & $54.7$ & $ {97.8}$ & $80.2$ \\
    MVDet \cite{hou2020multiview}                         & $88.2$ & $75.7$ & $94.7$ & $93.6$ & \quad & $83.9$ & $79.6$ & $96.8$ & $86.7$ \\
    MVM3Det \cite{haoran2021mvm3det}                        & $84.0$ & $75.8$ & $93.6$ & $90.2$ & \quad & $-$ & $-$ & $-$ & $-$  \\
    SHOT \cite{Song_2021_ICCV}                        & $ {90.2}$ & $ {76.5}$ & $ {96.1}$ & $ {94.0}$ & \quad & $ {88.3}$ & $ {82.0}$ & $96.6$ & $ {91.5}$  \\
    MVDeTr \cite{hou2021multiview}                         & $ {91.5}$ & $ {82.1}$ & $ {97.4}$ & $ {94.0}$ & \quad & $ {93.7}$ & $ {91.3}$ & $ {99.5}$ & $ {94.2}$ \\
    \midrule[0.4pt]
    VFA (ours) & $ {89.9}$ & $ {94.6}$ & $ {95.1}$ & $ {95.4}$ & \quad & $ {88.8}$ & $ {96.4}$ & $ {98.0}$ & $ {93.2}$ \\
    \bottomrule
    \end{tabular}
    }
\end{center}
\setlength{\abovecaptionskip}{0cm} 
\setlength{\belowcaptionskip}{+0.15cm}
\caption{Comparing with the state-of-the-art methods on Wildtrack and MultiviewX datasets for 2D multiview detection. We report and compare MODA, MODP, Precision and recall, all in percentage numbers. Higher is better.
}
\label{experiments_mx_wt}
\end{table*}

\section{Experiment}
\subsection{Experimental Setting}
In this section, we evaluate the performance of the proposed VFA on WildTrack\cite{8578626}, MultiviewX\cite{hou2020multiview}, MVM3D\cite{haoran2021mvm3det} and MultiviewC datasets.
\subsubsection{Datasets}
\textbf{Existing multiview detection datasets.} The \textbf{WildTrack} dataset is a real-world dataset while the \textbf{MultiviewX} dataset is a synthetic dataset for multiview pedestrian detection. Both datasets do not include the orientation label of the object. Therefore, the validation on these datasets is focus on occlusion cases.
The \textbf{MVM3D} for mobile robot detection is similar to our proposed dataset, containing images collected from multiple views, bounding boxes including position and orientation, and obstacles with different heights. However, it does not consider targets with diverse sizes. 
\begin{figure*}
\begin{center}
\includegraphics[width=1.0\linewidth]{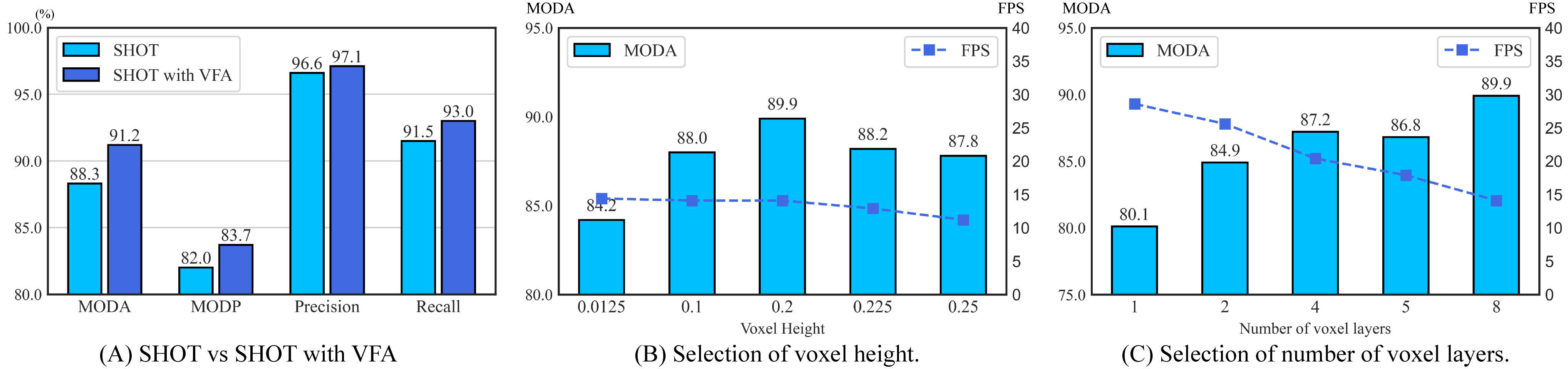}

\end{center}
\setlength{\abovecaptionskip}{0cm} 
\setlength{\belowcaptionskip}{+0.15cm}
\caption{\textbf{A}: Experimental comparison between the proposed VFA method and \cite{Song_2021_ICCV} on the MultiviewX dataset. After integrating VFA on top of \cite{Song_2021_ICCV}, we observe consistent improvement on all four metrics.
\textbf{B} and \textbf{C}: impact of hyper-parameters on MultiviewX, \emph{i.e.,} voxel height and the number of voxel layers, respectively. Results are reported on the MultiviewX validation set. When studying one hyper-parameter, we fix the others at a near optimal value. We report MODA (\%) and inference time (frames per second, FPS). We select voxel height and number of voxel layers as 0.2 and 8, respectively for this dataset. We conduct similar analysis on the validation set of each dataset. 
}
\label{H_NUM_vs_Perf_FPS}
\end{figure*}
\textbf{Our new MultiviewC dataset.} To address the need of more diverse and realistic test cases for multiview 3D detection, we present our new challenging dataset MultiviewC generated using Unreal Engine 4.26 \cite{unrealengine}. 
The scene is fixed on a cattle farm with sunny lighting conditions. MultiviewC dataset covers a square of 39 meters by 39 meters. We quantize the ground plane into a $3,900\times3,900$ grid. There are 7 cameras, with 4 located in the 4 corners of the area and 3 located on top of a drinking water trough. All cameras are calibrated, each of which captures every two seconds a $720\times 1,280$ pixel resolution image. We collect 560 images from each camera, so $3,920$ images in total for 7 cameras. Our interested targets are 15 cows with five random actions: sleeping, grazing, running, walking and lazing. To simulate more occlusion situations, the cows hold their current movement for one second when they encounter another cow. Different from existing datasets, the targets in MultiviewC have diverse sizes and aspect ratios. The height, width and length of each cow are distributed as follows: $1.1 \sim 1.5$ meters, $1.1 \sim 1.5$ meters and $2.48 \sim 2.78$ meters. Actions also affect their size, for example, a sleeping cow has lower height and longer length. The visualization of this dataset is shown in Fig. \ref{MultiviewC}. MultiviewC will be released on GitHub.

\subsubsection{Evaluation Metrics}
Following the existing methods such as \cite{haoran2021mvm3det, hou2020multiview, Baqu2017DeepOR}, we adopt Multi-Object Detection Precision (MODP) and Multi-Object Detection Accuracy (MODA), recall, precision as the evaluation metrics for the 2D object localization. For 3D detection, Average Precision of 3D detection ($A P_{3 D}$)\cite{6248074}, Average Orientation Similarity (AOS) and Orientation Score (OS) \cite{mousavian20173d} are selected.

\subsubsection{Implementation Details}
During experiments, input images of other sizes are resized to $720\times1,280$ pixel resolution to save memory. We employ a multi-scale feature extractor based on a ResNet-18 taking the output of the last three layers as the multi-scale feature. After feature extraction, before voxelized feature aggregation we adjust channel size of multi-scale features to 256. In VFA, size of voxel grid are dependent on plane size of scene and target's highest height. For example, the plane size of ground area in MultiviewC dataset is 39 meters by 39 meters, the height of cattle is from 1.1 meters to 1.5 meters, if we quantize planes into a $156\times156\times5$ grid and set the grid height to 1.6 meters, voxel size will be $0.25m\times0.25m\times0.32m$. During training, we use SGD optimizer with L2-normalization of $5E-4$ and momentum of 0.5. One-cycle learning rate scheduler\cite{smith2018superconvergence} with learning rate 0.02 is also adapted. We trained the network over 30 epochs with mini-batch size 1. All experiments were performed on RTX-3090 GPUs. 

\subsection{Evaluation of VFA}

\begin{table*}
\begin{center}
    \resizebox{0.99\linewidth}{!}{
    \begin{tabular}{l ccc ccc c lll lll }
    \toprule
    \multirow{3}{*}{Methods} & \multicolumn{6}{c}{MVM3D} & \quad & \multicolumn{6}{c}{MultiviewC} \\
    \cmidrule[0.4pt]{2-7} \cmidrule[0.4pt]{9-14}
    & \multicolumn{3}{c}{$\mathrm{IoU}=0.25$} & \multicolumn{3}{c}{$\mathrm{IoU}=0.5$} & \quad & \multicolumn{3}{c}{$\mathrm{IoU}=0.25$} &
    \multicolumn{3}{c}{$\mathrm{IoU}=0.5$} \\
    \cmidrule[0.4pt]{2-7} \cmidrule[0.4pt]{9-14}
    & $A P_{3 D}$ & AOS & OS & $A P_{3 D}$ & AOS & OS & \quad  & $A P_{3 D}$ & AOS & OS & $A P_{3 D}$ & AOS & OS \\
    \midrule[0.4pt]
    Deep3DBox \cite{mousavian20173d} & $77.2\%$ & $67.1 \%$ & $0.87$ & $30.3 \%$ & $26.6 \%$ & $0.88$ & \quad & $84.2 \%^\ast$ & $75.9 \%^\ast$ & $0.90^\ast$ & $33.6 \%^\ast$ & $30.4 \%^\ast$ & $0.90^\ast$ \\
    MVM3Det \cite{haoran2021mvm3det} & $90.2 \%$ & $\mathbf{82.6} \%$ & $\mathbf{0.91}$ & $49.0 \%$ & $45.5 \%$ & $\mathbf{0.92}$ & \quad & $92.2 \%^\ast$ & $85.7 \%^\ast$ & $0.93^\ast$ & $55.6 \%^\ast$ & $51.5 \%^\ast$ & $0.92^\ast$ \\
    \midrule[0.4pt]
    VFA (w/ PD) & $-$ & $-$ & $-$ & $-$ & $-$ & $-$ & \quad & $95.5\%$ & $95.1\%$ & $0.99$ & $89.2\%$ & $88.7\%$ & $0.99$ \\
    VFA (w/ GD) & $-$ & $-$ & $-$ & $-$ & $-$ & $-$ & \quad & $95.9\%$ & $95.0\%$ & $\mathbf{0.99}$ & $89.4\%$ & $88.8\%$ & $0.99$ \\
    VFA (w/ OGD) & $\mathbf{90.9}\%$ & $81.7\%$  & $0.89$ & $\mathbf{88.7} \%$ & $\mathbf{77.6} \%$ & $0.87$ & \quad & $\mathbf{96.5}\%$ & $\mathbf{95.2}\%$ & $0.98$ & $\mathbf{89.8}\%$ & $\mathbf{88.8}\%$ & $\mathbf{0.99}$ \\
    \bottomrule
    \end{tabular}
    }
\end{center}
\setlength{\abovecaptionskip}{0cm} 
\setlength{\belowcaptionskip}{+0.15cm}
\caption{
    Method comparison on MVM3D and MultiviewC datasets for 3D multiview detection. Apart from comparing with two recent approaches \cite{mousavian20173d,haoran2021mvm3det}, we also evaluate our system with different ground-truth modeling schemes: point distribution (w/ PD), spherical Gaussian distribution (w/ GD) and oriented Gaussian distribution (w/ OGD). $^\ast$ indicate results obtained through our re-implementation on the MultiviewC dataset.
}
\label{experiments_mvm3d_mc}
\end{table*}

\begin{figure*}
\begin{center}
\includegraphics[width=0.99\linewidth]{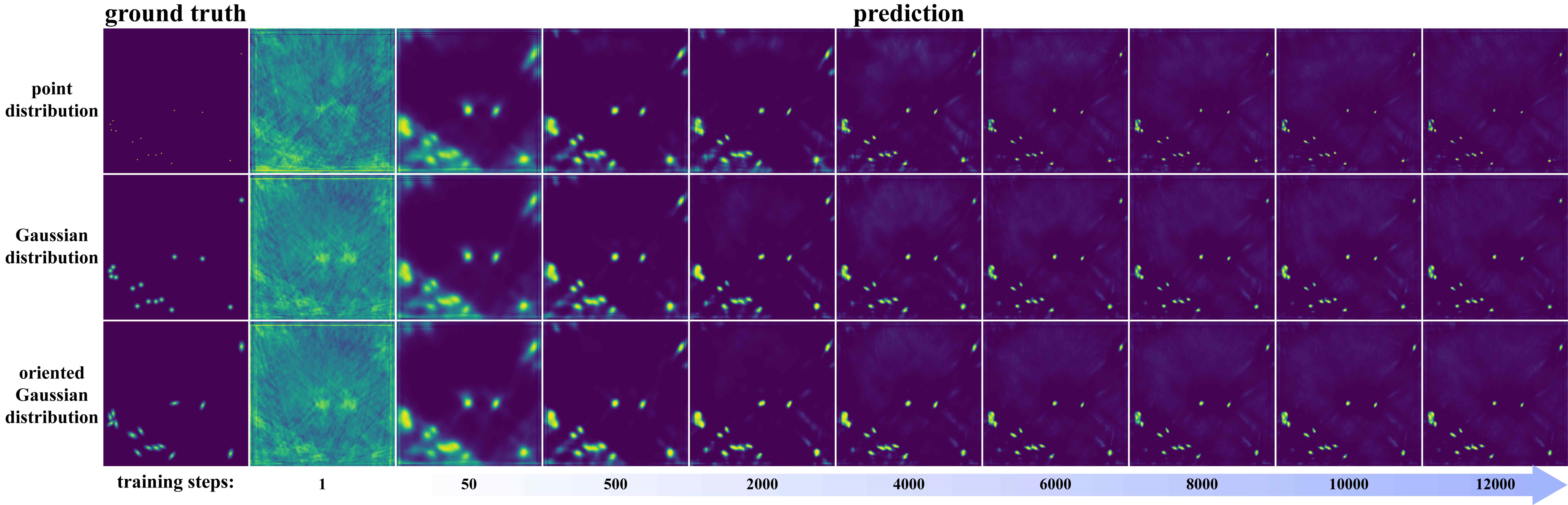}

\end{center}
\caption{Visualization of network predictions along the training iterations. The three ground-truth modeling methods are used in Row 1, Row 2, and Row 3, respectively. The detection outputs improve as training goes.  
}
\label{RGD_GD_GP}
\end{figure*}



\textbf{Comparison with the state of the art.}
Results on \textbf{WildTrack} dataset and \textbf{MultiviewX} dataset: the evaluation is focused on how our proposed method deals with occlusion cases. Comparison between our VFA method and previous methods on multiview 2D detection is shown in Table \ref{experiments_mx_wt}. Our VFA method outperforms the previous state-of-the-art method MVDeTr by $+12.5\%$ MODP, $+1.4\%$ Recall on the Wildtrack dataset, and $+5.1\%$ MODP on the MultiviewX dataset and the results of other metrics are close. MVDeTr method utilizes transformer to recover projection distortion while we apply voxelized feature aggregation to reduce the projection distortion achieving similar competitive result. On \textbf{MVM3D} dataset and \textbf{MultiviewC} dataset, we focus on 3D detection, including position and orientation. The comparison between our proposed VFA method with MVM3Det method and a single view method, Deep3DBox\cite{mousavian20173d} is shown in Table \ref{experiments_mvm3d_mc}. Our VFA method is trained with 3 different encoding distributions: point distribution (PD), Gaussian distribution (GD), and oriented Gaussian distribution (OGD). The results show that although on par on MVM3D dataset, our VFA method outperforms Deep3DBox and MVM3Det by large margins on all metrics on MultiviewC dataset. 
The reason is that MultiviewC dataset contains cow elongated bodies and different poses such as standing and lying, therefore more challenging than MVM3D dataset. In case of encoding methods, our proposed OGD encoding can further enhance performance. 



\textbf{Comparison with \cite{Song_2021_ICCV}.} The SHOT method \cite{Song_2021_ICCV} is the closest work to ours, which also addresses the projection distortion problem. To demonstrate the effectiveness of our proposed VFA, we change SHOT by replacing its 2D-3D feature projection by VFA component and call this as ``SHOT with VFA''. We then compare SHOT and ``SHOT with VFA'' under the same experiment settings on the MultiviewX dataset. Fig. \ref{H_NUM_vs_Perf_FPS}  {A} shows that the ``SHOT with VFA'' method outperforms the original SHOT on all metrics. 

\textbf{Hyper-parameter analysis.}\label{voxel_grids_experiment}
There are two primary hyper-parameters of VFA which are \textbf{grid height} and \textbf{number of voxel layers}.
First, we assess the influence of \textbf{grid height} as shown in Fig.\ \ref{H_NUM_vs_Perf_FPS}  {B}  (and Table 4 in the Supplementary Material). It is clear that our VFA performs best with the height of 1.6 meters and the grids size of $0.1m\times0.1m\times0.2m$ for both Wildtrack and MultiviewX datasets. We then find the optimal \textbf{number of voxel layers} while the grid height is fixed to 1.6 meters as shown in the Fig. \ref{H_NUM_vs_Perf_FPS}  {C}  (and in Table 4 of Supplementary Material). Results show that more layers achieve better performance as well as higher computational complexity, and 8 layers seems to be a good balance.

\subsection{Further Analysis}
\textbf{Influence of different number of input views.} 
\begin{table}
\begin{center}
    \small 
    \resizebox{0.99\linewidth}{!}{
    \begin{tabular}{cccc c ccc}
    \toprule 
    \multirow{2}{30pt}{Num of Views } & \multicolumn{3}{c}{$\mathrm{IoU}=0.25$} & \quad & \multicolumn{3}{c}{ IoU $=0.5$} \\ 
    & $A P_{3 D}$ & AOS & OS & \quad & $A P_{3 D}$ & AOS & OS \\ 
    \midrule[0.4pt]
    1  & $81.5\%$ & $79.6\%$ & $0.98$ & \quad & $68.4\%$ & $67.4\%$ & $0.99$ \\
    2  & $90.8\%$ & $89.3\%$ & $0.98$ & \quad & $89.0\%$ & $87.9\%$ & $0.99$ \\
    4  & $90.9\%$ & $89.8\%$ & $\mathbf{0.99}$ & \quad & $89.9\%$ & $\mathbf{89.8}\%$ & $0.99$ \\
    7 & $\mathbf{96.5}\%$ & $\mathbf{95.2}\%$ & $0.98$ & \quad & $\mathbf{90.0}\%$ & $ 89.3\%$ & $\mathbf{0.99}$ \\
    \bottomrule
    \end{tabular}
    }
\end{center}
\setlength{\abovecaptionskip}{0cm} 
\setlength{\belowcaptionskip}{+0.15cm}
\caption{
    Impact of using different number of camera views on the MultiviewC dataset. Three metrics under two IoU values are compared. The proposed VFA method is used. Clearly using more views offers us more higher detection accuracy. 
}
\label{ablation_mc}
\end{table}

\begin{table}
\begin{center}
    \small 
    \resizebox{0.99\linewidth}{!}{
    \begin{tabular}{cccc c ccc}
    \toprule 
    \multirow{2}{30pt}{Num of Views } & \multicolumn{3}{c}{$\mathrm{IoU}=0.25$} & \quad & \multicolumn{3}{c}{ IoU $=0.5$} \\ 
    & $A P_{3 D}$ & AOS & OS & \quad & $A P_{3 D}$ & AOS & OS \\ 
    \midrule[0.4pt]
    1  & $81.5\%$ & $79.6\%$ & $0.98$ & \quad & $68.4\%$ & $67.4\%$ & $0.99$ \\
    2  & $90.8\%$ & $89.3\%$ & $0.98$ & \quad & $89.0\%$ & $87.9\%$ & $0.99$ \\
    4  & $90.9\%$ & $89.8\%$ & $\mathbf{0.99}$ & \quad & $89.9\%$ & $\mathbf{89.8}\%$ & $0.99$ \\
    7 & $\mathbf{96.5}\%$ & $\mathbf{95.2}\%$ & $0.98$ & \quad & $\mathbf{90.0}\%$ & $ 89.3\%$ & $\mathbf{0.99}$ \\
    \bottomrule
    \end{tabular}
    }
\end{center}
\setlength{\abovecaptionskip}{0cm} 
\setlength{\belowcaptionskip}{+0.15cm}
\caption{
    Impact of using different number of camera views on the MultiviewC dataset. Three metrics under two IoU values are compared. The proposed VFA method is used. Clearly using more views offers us more higher detection accuracy. 
}
\label{ablation_mc}
\end{table}

We evaluate the impact of view number. 
To be specific, we experiment on 1, 2, 4 and 7 camera views among which only the four cameras at the corners are used in the first three cases.

As shown in Table \ref{ablation_mc}, our method performs increasingly better from single view to multiple views indicating that VFA is truly capable of not only working with a single view but also aggregating multiview feature and alleviating occlusion problem.
However, it differs slightly between two and four views which is because the additional two corner cameras can provide limited useful multiview features for detection.  
While using all camera views, the performance is further improved. We consider that the remaining three cameras above the trough contain additional multiview cues.


\begin{figure}
\begin{center}
\includegraphics[width=0.9\linewidth]{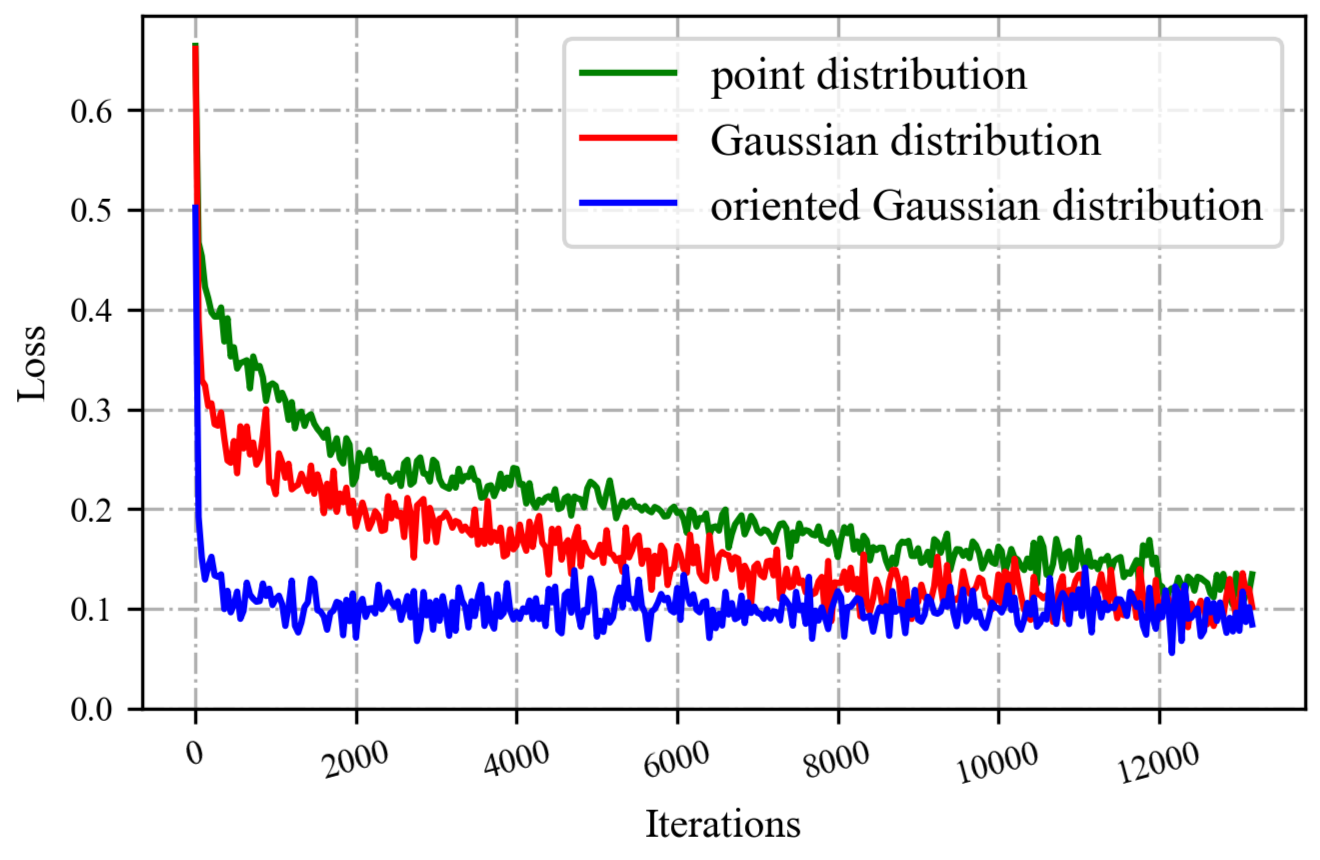}

\end{center}
\caption{
Training loss curves (loss value \emph{vs.} training iterations) of different ground-truth modeling methods. 
Oriented Gaussian distribution converges much faster than the other methods.
}
\label{Loss_RGD_GD_GP}
\end{figure}

\textbf{Effectiveness of oriented Gaussian distribution.}\label{Effectiveness of RGD}
Table \ref{experiments_mvm3d_mc} shows that VFA using oriented Gaussian distribution achieves better precision than using point and Gaussian distributions.
In addition, we visualize in the Fig. \ref{RGD_GD_GP} the evolutions of predicted heat maps from three encoding distributions at different training steps. 
We notice that all initial predictions tend to share a similar pattern that strong blobs are oriented and naturally aligned with the direction of the targets. Therefore, it is sensible that the orientation should be encoded in the aggregated features and predictions. Thus, instead of point distribution or Gaussian distribution, we train our VFA network using labels encoded by oriented Gaussian distribution. 
Finally, as shown in the Fig. \ref{Loss_RGD_GD_GP}, although VFA with each of the 3 encoding distributions can reach comparable losses after enough training, oriented Gaussian distribution converges extremely fast, at least $4X$ faster than the others.

\section{Conclusions}

This paper proposes the voxelized 3D feature aggregation (VFA) method to resolve the problem of projection distortion caused by the usage of 2D transformation schemes. Our method, in a straightforward way, maps Z-axis-aware 3D voxels onto the 2D multi-view images, allowing features along the same vertical line to be projected onto the same position on the ground plane. VFA purifies the aggregated features on the ground plane and thus improves the system accuracy. On a side contribution, we use an oriented Gaussian modeling method for elongated objects on the ground plane, like cattle. This method allows for more efficient training process and slightly better accuracy. We validate this system on four datasets (including a newly introduced cattle detection dataset MultiviewC), where we report very competitive detection accuracy compared with the state-of-the-art system.

\section*{Acknowledgments}
We thank Liang Zheng for his comments on the text and constructive discussions. This work is partially funded by CSIRO's Julius Career Award (JCA20-899) and Automated Farm Provenance.

\small
\bibliographystyle{IEEEtran}
\bibliography{vfatip}

\newpage

 
\vspace{-30pt}

\begin{IEEEbiography}[{\includegraphics[width=1in,height=1.25in,clip,keepaspectratio]{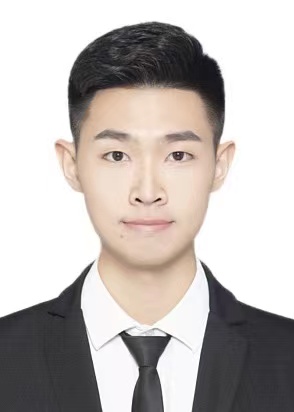}}]{Jiahao Ma }
is a Master student with College of Engineering and Computer Science, Australian National University. His current research interests falls in multiview detection and novel view synthesis.
\end{IEEEbiography}

\begin{IEEEbiography}[{\includegraphics[width=1in,height=1.25in,clip,keepaspectratio]{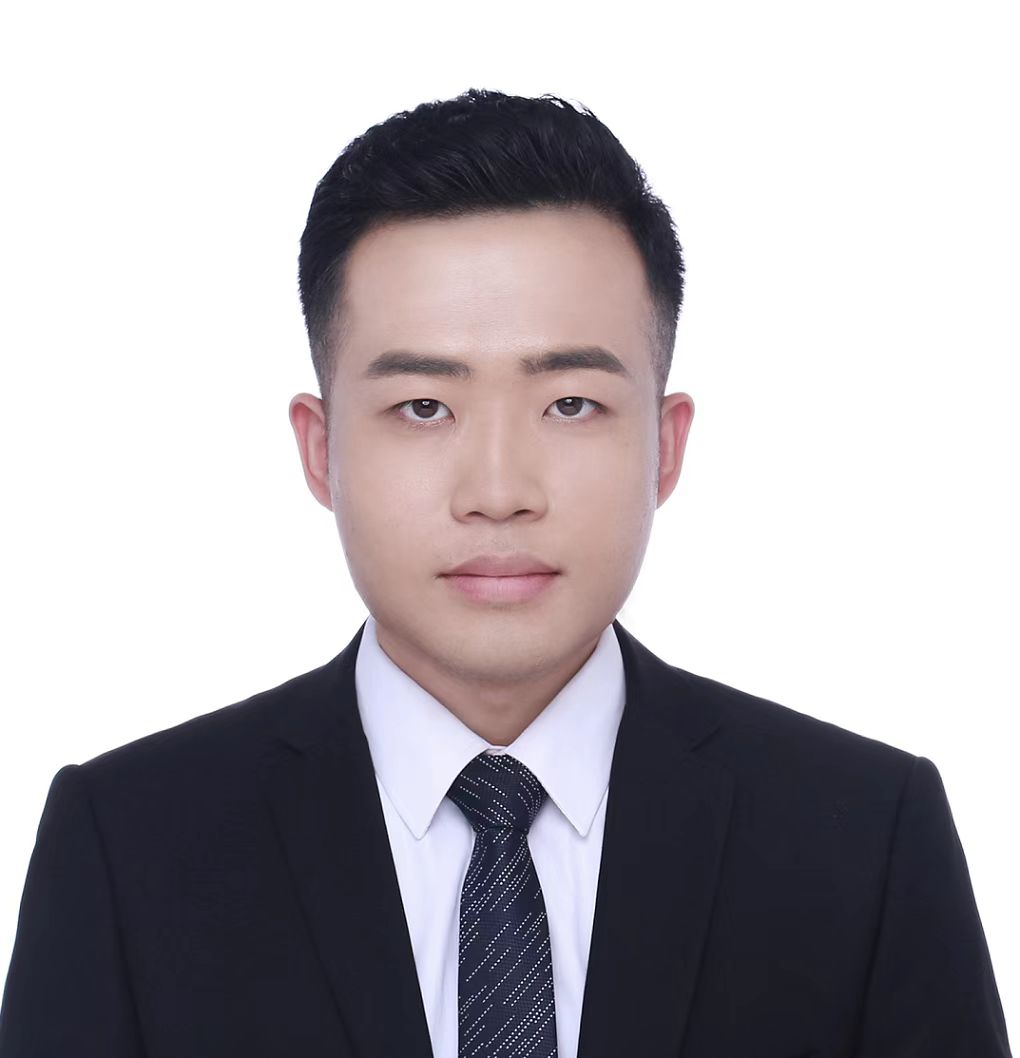}}]{Jinguang tong }
is currently a Ph.D. student with the College of Engineering and Computer Science, Australian National University. He received the Master degree in Optical Engineering from Zhejiang University, China, in 2015. His current research interest falls in 3D computer vision and deep learning.
\end{IEEEbiography}

\begin{IEEEbiography}[{\includegraphics[width=1in,height=1.25in,clip,keepaspectratio]{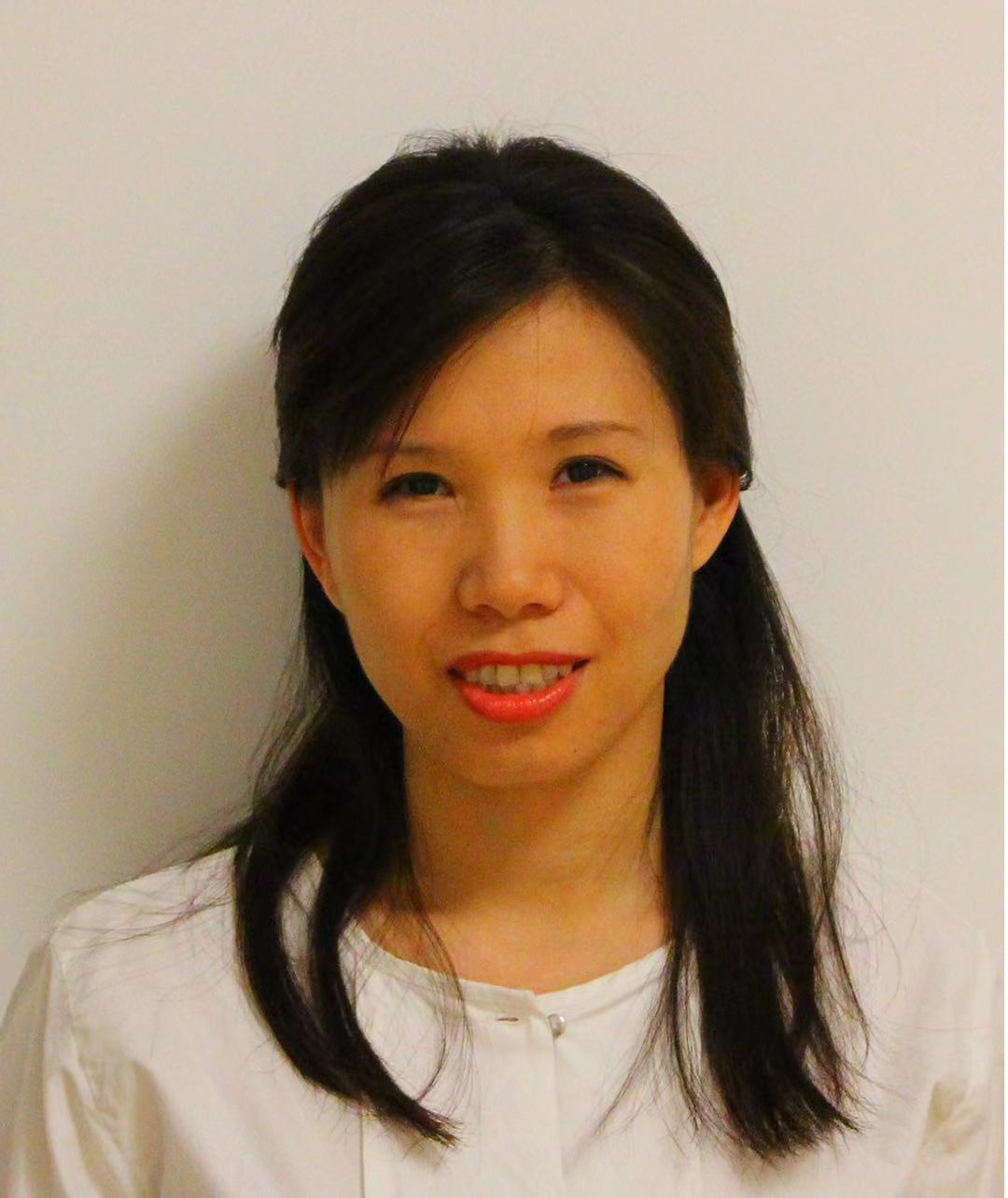}}]{Shan Wang }
is currently a PhD student with Research School of Computing, the Australian National University, DATA61-CSIRO (Canberra, Australia). Her main research interests include localization, 3D Reconstruction, Event camera.
\end{IEEEbiography}

\begin{IEEEbiography}[{\includegraphics[width=1in,height=1.25in,clip,keepaspectratio]{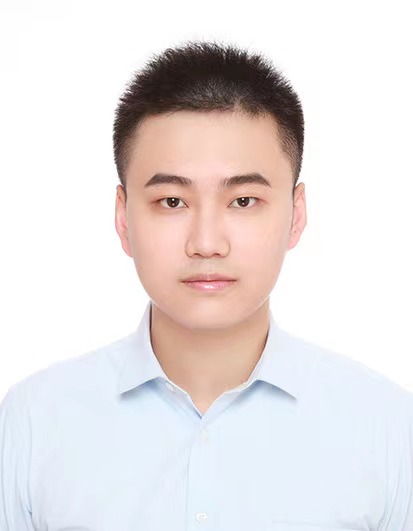}}]{Zicheng Duan }
is a Master student with College of Engineering and Computer Science, Australian National University. His current research interests include multiview detection and 3D reconstruction.
\end{IEEEbiography}

\begin{IEEEbiography}[{\includegraphics[width=1in,height=1.25in,clip,keepaspectratio]{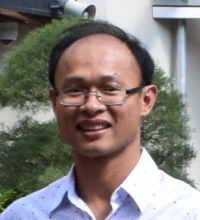}}]{Chuong Nguyen }
received his PhD degree in Engineering from Monash University in 2010. Then he spent 3 years as a research fellow at CSIRO in Australia working on 3D reconstruction methods for insects and plants. He is now a senior research scientist at CSIRO Data61 leading initiatives in 3D digitisation, health \& animal monitoring, environmental survey, and advanced manufacturing. His research interests include 3D computer vision, machine learning and robotics.
\end{IEEEbiography}

\vspace{11pt}



\end{document}